\documentclass[letterpaper]{article} 
\usepackage[preprint]{aaai2027}  
\usepackage[hyphens]{url}  
\usepackage{graphicx} 
\usepackage{natbib}  
\usepackage{caption} 
\usepackage{algorithm}
\usepackage{algorithmic}

\usepackage{newfloat}
\usepackage{listings}
\DeclareCaptionStyle{ruled}{labelfont=normalfont,labelsep=colon,strut=off} 
\floatstyle{ruled}
\newfloat{listing}{tb}{lst}{}
\floatname{listing}{Listing}

\usepackage{booktabs}
\usepackage{enumitem} 
\usepackage{booktabs} 
\usepackage{multirow} 
\usepackage{amsmath} 
\usepackage{tabularx} 
\usepackage{microtype} 
\usepackage{xcolor} 
\usepackage{tcolorbox} 
\usepackage{makecell}
\usepackage{array} 
\usepackage[table]{xcolor}

\newcommand{\best}[1]{\textbf{#1}}
\newcommand{\gb}[1]{\underline{#1}}
\newcommand{\meth}[1]{\hspace{1.2em}#1}

\newcolumntype{M}{>{\raggedright\arraybackslash}X}
\newcommand{\meanstd}[2]{#1\,{\scriptsize$\pm$\,#2}}
\title{Schema-Aware Planning with Hybrid Knowledge Toolset for Knowledge Graph Triple Verification}
\author{
Xinyan Ma\textsuperscript{\rm 1},
Xianhao Ou\textsuperscript{\rm 1},
Chang Li\textsuperscript{\rm 1},
Yexing Du\textsuperscript{\rm 1},
Zhihao Zhu\textsuperscript{\rm 1},
Shixin Jiang\textsuperscript{\rm 1},
Runxuan Liu\textsuperscript{\rm 1},
Dandan Tu\textsuperscript{\rm 2},
Ming Liu\textsuperscript{\rm 1}\thanks{Corresponding author.}
}

\affiliations{
\textsuperscript{\rm 1}Harbin Institute of Technology, Harbin, China\\
\textsuperscript{\rm 2}Huawei Technologies Co., Ltd., Beijing, China
}

\begin{document}

\maketitle

\begin{abstract}
Knowledge Graphs (KGs) provide a critical foundation for many AI systems, yet their automated construction inevitably introduces noisy triples that undermine data trustworthiness. Reliable KG auditing requires jointly examining internal structural constraints and open-world evidence, whereas existing methods often rely on a single evidence source or static, one-shot inference, limiting adaptive evidence acquisition and auditability. We propose SHARP (Schema-Hybrid Agent for Reliable Prediction), a training-free agent that reformulates triple verification as a dynamic process of strategic planning, active investigation, and evidential reasoning. SHARP uses retrieved historical verification trajectories and entity--relation schema information to construct a task-specific investigation plan. It then iteratively coordinates KG and textual tools to cross-validate structural and semantic evidence, producing a verification verdict with a transparent evidence chain. On independently sampled, balanced test sets constructed from FB15K-237 and Wikidata5M-Ind, SHARP achieves average accuracies of 87.1\% and 93.6\%, respectively. Controlled comparisons with standard RAG using the same evidence sources, together with ablation studies, further show that the gains arise not merely from external retrieval, but from the combined effects of schema-aware planning, memory augmentation, and iterative hybrid tool use.

\end{abstract}


\section{Introduction}

Knowledge Graphs (KGs) represent knowledge as structured triples $(h, r, t)$ and have been widely used in question answering~\cite{Intro_Complex}, recommender systems~\cite{Intro_KG_recommend}, search engines, and hallucination mitigation for large language models~\cite{Intro_can,Intro_ontology}. Yet automatically extracted or integrated graphs inevitably contain wrong mappings, stale values, relation-scope errors, and unsupported edges. Auditing a candidate triple is therefore not only a compatibility-scoring problem: a useful auditor must resolve the identities and relation schema, locate relevant evidence, reconcile structural and textual signals, and expose what supports its verdict.

\begin{figure}[t]
    \centering
    \includegraphics[width=\linewidth]{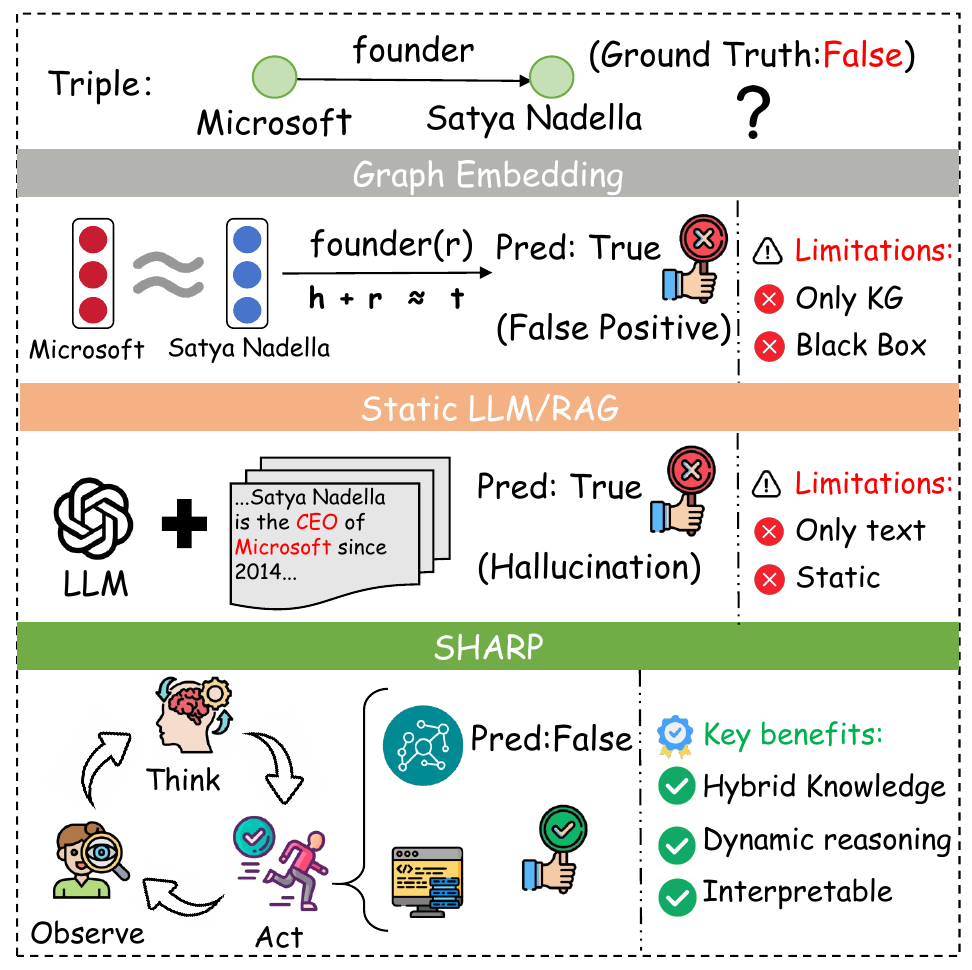} 
    \caption{Comparisons between previous methods and our proposed SHARP. Existing graph embedding and static LLM/RAG methods may produce false positives, while SHARP integrates structural and textual evidence through dynamic reasoning to detect errors.}
    \label{FIG:intro}
\end{figure}

Existing approaches offer complementary but incomplete views. Embedding models such as TransE and RotatE learn structural compatibility~\cite{Intro_TransE,Intro_RotatE}; textual encoders such as KG-BERT and SimKGC enrich triples with descriptions~\cite{Intro_KG_BERT,Re_SimKGC}. These fixed scorers do not actively collect candidate-specific evidence. LLM-based retrieval can consult open text~\cite{Intro_fact}, but a static batch cannot revise its investigation after finding an alias, temporal qualifier, or relation-direction mismatch. KG reasoning agents can explore graph paths~\cite{Intro_think_on_graph,Re_PlanOnGraph}, but primarily target question answering rather than binary auditing with graph and open-world evidence. Figure~\ref{FIG:intro} illustrates how SHARP uses active investigation and evidential reasoning to reject an erroneous triple that similarity-based or static LLM/RAG methods may accept.

We introduce SHARP, a training-free agent that formulates binary KG triple verification as a plan-retrieve-reason process over relation schemas, retrieved verification trajectories, and hybrid KG-text evidence. We reformulate triple verification from a traditional classification task into a human-like cognitive process of ``strategic planning – dynamic investigation – evidential reasoning.''
Specifically, SHARP emulates the workflow of a human fact-checker. First, using a memory-augmented mechanism, it performs analogical reasoning over similar historical verification trajectories and generates a global plan tailored to the schema and characteristics of the target triple. It then enters a \textit{Think-Act-Observe} loop, leveraging a hybrid toolset that accesses both internal KG structures and external knowledge to dynamically collect evidence and iteratively refine reasoning. This mechanism bridges information silos, achieving complementarity between structured KG constraints and unstructured semantic evidence, and outputs both a verification result and an inspectable evidence trace.

We evaluate SHARP on two KG verification benchmarks through comparisons with representative baselines, controlled comparisons with standard RAG, and component ablation studies. SHARP consistently outperforms existing methods, while controlled experiments demonstrate that the gains go beyond retrieval augmentation and arise from schema-aware planning, memory augmentation, and hybrid evidence integration.

In summary, our main contributions are as follows:
\begin{itemize}
    \item We formulate KG triple verification as an interactive plan-retrieve-reason process and instantiate it with SHARP, which couples relation-schema planning with retrieved, manually refined verification trajectories.
    \item We implement a five-tool interface that exposes entity and relation definitions, local KG neighborhoods, short relational paths, Wikipedia evidence, and web snippets within one iterative evidence-acquisition loop.
    \item  We evaluate SHARP on three independently sampled test sets per benchmark. Same-source RAG comparisons across three backbones, component removals, evidence-trace audits, and cost measurements characterize both its gains and its limitations under an explicit information boundary.
\end{itemize}


\begin{figure*}[htbp]
    \centering
    \includegraphics[width=\linewidth]{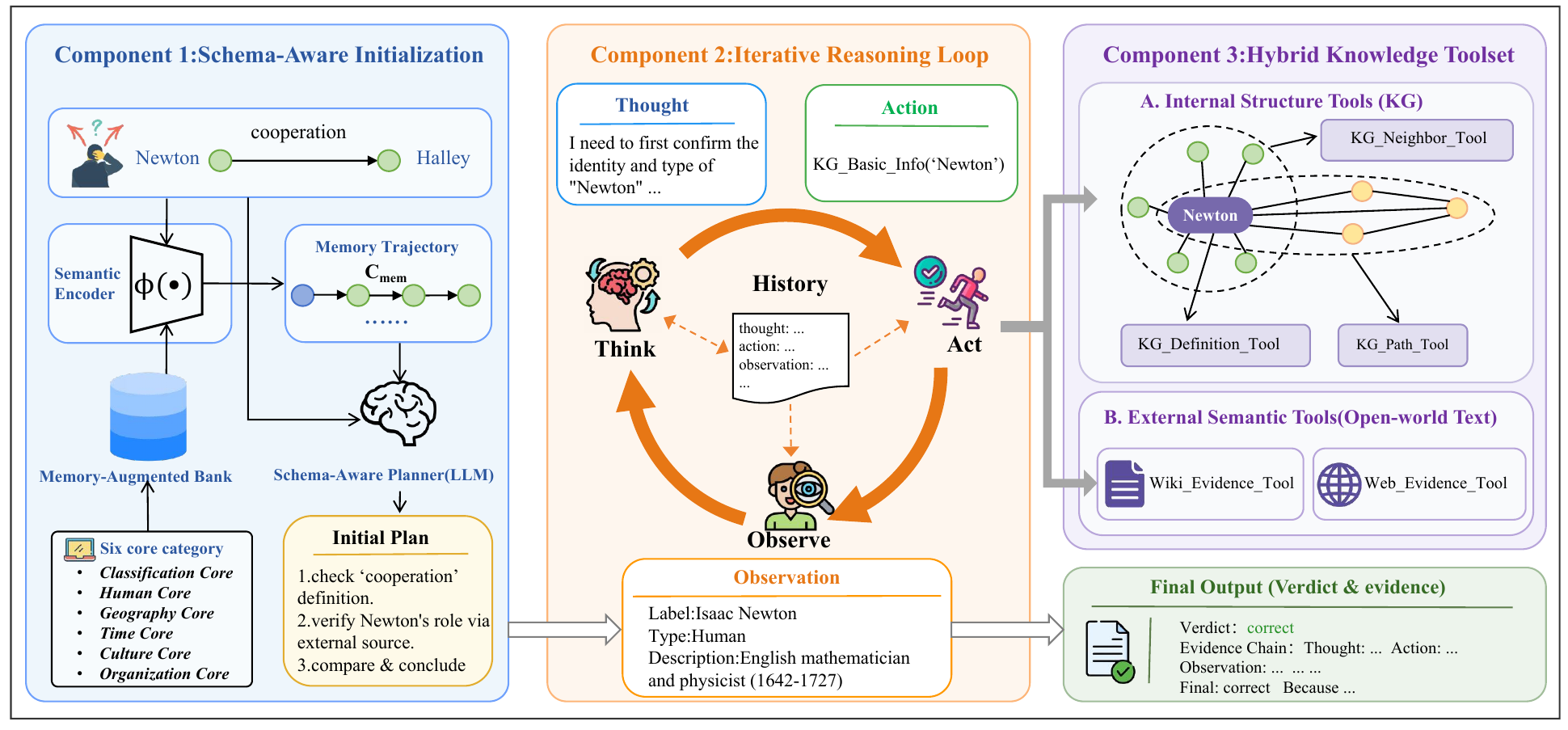}    
    \caption{Overview of the SHARP architecture. The framework consists of three components: (1) Schema-Aware Initialization, which retrieves similar trajectories from memory to generate an initial plan; (2) an Iterative Reasoning Loop, which dynamically updates reasoning through an improved ReAct cycle; and (3) a Hybrid Knowledge Toolset, which combines internal KG and external text tools for cross-validated verification.}
    \label{FIG:method}
\end{figure*}

\section{Related Work}

\paragraph{KG completion and verification.}
Path- and rule-based methods infer relations from recurring graph structures~\cite{Re_PRA,Re_AMIE}. Graph embedding models, including TransE~\cite{Intro_TransE}, DistMult~\cite{Re_DistMult}, and RotatE~\cite{Intro_RotatE}, learn structural representations to capture relational patterns. PLM-based methods, such as KG-BERT~\cite{Intro_KG_BERT}, KEPLER~\cite{Re_KEPLER}, and SimKGC~\cite{Re_SimKGC}, incorporate textual semantics, while LLM-based approaches use retrieval-augmented evidence for traceable verification~\cite{Re_MPIKGC}. Existing approaches rely on fixed structural or textual evidence; SHARP instead integrates relation schemas, graph structure, and open-world evidence.

\paragraph{LLM-based Fact-Checking.}
LLM-based methods extend fact-checking paradigms from different perspectives. SelfCheckGPT~\cite{Re_SelfCheckGPT} uses sampling-based self-consistency, FActScore~\cite{Re_FActScore} evaluates atomic facts, and RARR~\cite{Re_RARR}, Verify-and-Edit~\cite{Re_Verify-and-Edit}, and CoVe~\cite{Re_CoVe} revise outputs through retrieval, verification, or self-checking. KG-enhanced methods such as KGV~\cite{Re_KGV} and R3~\cite{Re_R3} ground verification in semantic graphs or KG triples. SHARP fills this gap with candidate-specific schema and path probing for binary KG triple verification.

\paragraph{Autonomous Agents.}
Autonomous agents equip LLMs with planning, action, reflection, and collaboration. ReAct~\cite{Re_React} couples reasoning with action, Toolformer~\cite{Re_Toolformer} enables tool use, Reflexion~\cite{Re_Reflextion} supports self-correction, and AutoGen~\cite{Re_AutoGen} and MetaGPT~\cite{Re_MetaGPT} enable multi-agent collaboration. For verification, FactAgent~\cite{Re_Factagent} connects claim analysis, evidence search, and verdict aggregation, while SAFE~\cite{Re_SAFE} verifies atomic facts with retrieval-based agents. KG-Agent~\cite{Re_KGagent} and MAKGED~\cite{Re_MAKGED} apply agents to KG reasoning and error detection. SHARP adapts agentic reasoning to KG auditing through schema-aware planning, retrieved verification trajectories, and hybrid evidence acquisition.

\section{SHARP}
In this section, we first provide an overview of the overall architecture of SHARP. Subsequently, we will delve into the three core components of this framework respectively: Schema-Aware Initialization, Iterative Reasoning Loop, and Hybrid Knowledge Toolset.

\subsection{Framework Overview}

We formulate the triple verification task as an interactive sequential reasoning process based on hybrid knowledge retrieval. As illustrated in Figure~\ref{FIG:method}, SHARP is built upon large language models (LLMs) and adopts the Think-Act-Observe ReAct paradigm. To address the cold-start issue and the separation between structural and semantic reasoning in complex logic handling, SHARP decomposes its reasoning process into three components: 
(1) \textbf{Schema-Aware Initialization:} The agent analyzes the structural and semantic features of the query triple $\tau_q$, retrieves similar historical reasoning trajectories via a memory-augmented mechanism, and combines them with the graph schema to generate an initial strategy $P_{init}$ as high-level heuristic guidance for subsequent investigation. 
(2) \textbf{Iterative Reasoning Loop:} Guided by $P_{init}$, the agent executes a dynamic ReAct loop, continuously refining the investigation strategy and verifying evidence through repeated Think, Act, and Observe steps. 
(3) \textbf{Hybrid Knowledge Toolset:} This module decomposes knowledge acquisition into internal structural probing and external semantic verification, and designs a cross-validation toolset with five atomic capabilities accordingly.

\subsection{Schema-Aware Initialization}
\label{ssec:Schema-Aware Initialization}

\subsubsection{Memory-Augmented Mechanism}
\label{sec:memory-schemawise}
To equip the agent with expert-like reasoning patterns and maintain consistent guidance, we replace static few-shot prompting with a global memory-augmented mechanism. We construct an external memory bank 
$\mathcal{M} = \{(x_i, \mathrm{traj}_i)\}_{i=1}^{N}$,
where initial trajectories are first generated through agent self-exploration and then manually refined. To cover heterogeneous relation schemas in the memory bank, we group high-frequency Wikidata relations into six categories: \textbf{Classification, Human, Geography, Time, Culture,} and \textbf{Organization}. To ensure robustness and generalization, high-frequency Wikidata triples are grouped into six core categories: 

The memory bank contains 200 manually refined verification trajectories covering 50 high-frequency relations across these categories. Each trajectory $\mathrm{traj}_i$ records the full reasoning process for a verification task $x_i=(h_i,r_i,t_i)$ as a Think--Act--Observe sequence (examples in the Appendix).

For a query triple $\tau_q=(h_q,r_q,t_q)$, a semantic encoder $\phi(\cdot)$ retrieves the top-$k$ most similar trajectories from $\mathcal{M}$ to form the contextual memory set $\mathcal{C}_{mem}$:
\begin{equation*}
    \text{sim}(\tau_q, x_i) = \frac{\phi(\tau_q) \cdot \phi(x_i)}{\|\phi(\tau_q)\| \, \|\phi(x_i)\|}
\end{equation*}
\begin{equation*}
    \mathcal{C}_{mem} = \bigcup_{j=1}^k \left\{ \mathrm{traj}_j \;\middle|\; x_j \in \mathop{\arg\max}_{x_i \in \mathcal{M}}^{(k)} \text{sim}(\tau_q, x_i) \right\}
\end{equation*}

Importantly, $\mathcal{C}_{mem}$ serves as a persistent reference in the agent’s working memory rather than just a cold-start prompt. During reasoning, the agent continuously consults these trajectories to reuse task-specific investigation patterns and tool-selection examples throughout the reasoning process.

\subsubsection{Schema-Aware Strategic Planning}
To avoid unguided exploration over complex graphs, the agent first performs a metacognitive planning step. Using a customized prompt, it integrates knowledge of the toolset $\mathcal{T}$, the graph schema features $\mathcal{S}$ of $\tau_q$, and the retrieved memory $\mathcal{C}_{mem}$. The LLM policy $\pi_\theta$ then generates an initial verification plan $P_{init}$:
\begin{equation*}
    P_{init} = \pi_\theta \big( \operatorname{Prompt}_{plan}(\tau_q, \mathcal{S}, \mathcal{C}_{mem}, \mathcal{T}) \big)
\end{equation*}

This plan specifies structured verification steps such as checking entity definitions and type constraints, retrieving 1-hop neighbors to verify direct relations, and consulting external sources to validate event chronology. Finally, $P_{init}$ is injected as the initial observation $\mathrm{Obs}_0$ into the agent's initial context $\mathcal{C}_0$, providing global guidance and serving as a persistent heuristic anchor for subsequent tool selection.

\subsection{Reasoning with ReAct Paradigm}
\label{ssec:Reasoning with ReAct Paradigm}

To better adapt to the KG triple verification task, we enhance the standard ReAct paradigm with a Plan Adherence and Correction Mechanism. The reasoning process is modeled as a dynamic discrete-time sequence $t = 1, 2, \dots, T$.

At step $t$, the agent operates on the global  reasoning context 
$\mathcal{H}_t = (I, \mathcal{C}_{mem}, \mathcal{P}_{init}, h_{0:t-1})$,
comprising system instructions $I$, retrieved expert demonstrations $\mathcal{C}_{mem}$, the initial strategic plan $\mathcal{P}_{init}$, and accumulated interaction history. As outlined in Algorithm~\ref{alg:1}, the agent  iteratively executes three recursive sub-processes:

\paragraph{Think:}
The agent derives the next reasoning strategy $Th_t$ using $\mathcal{P}_{init}$ and accumulated observations $Obs$. This ensures adherence to the original task goal while dynamically refining the plan based on newly observed evidence.

\paragraph{Act:}
Guided by $th_t$, the agent selects the optimal tool $a$ from the toolset $\mathcal{T}$ and generates execution arguments $p$:
\begin{equation*}
ac_t = (a, p) \sim \pi_\theta(ac_t \mid \mathcal{H}_t, th_t),
\end{equation*}
where $\pi_\theta$ is the action policy distribution parameterized by the LLM.

\paragraph{Observe:}
The environment (i.e., the Knowledge Graph interface or external search engine) executes $ac_t$, producing $obs_t = Env(ac_t)$. The new observation updates the global context to $\mathcal{H}_{t+1}$ and is appended to the interaction history.

The loop continues until the agent issues a termination action \texttt{Finish[Answer]} or reaches the maximum step $T_{max}$. To prevent infinite loops, a Mandatory Judgment Mechanism triggers at $T_{max}$, compelling the agent to perform probabilistic reasoning over available information to produce a final verdict. This design ensures effective coordination between structured macro-planning guided by $\mathcal{P}_{init}$ and dynamic micro-execution driven by ReAct. Prompt templates for the cyclic reasoning and forced judgment phases are in the Appendix. 

\begin{algorithm}[t]
\caption{SHARP Reasoning Algorithm}
\label{alg:1}
\begin{algorithmic}[1]

\REQUIRE Target Triple $\tau$;
         System Instruction $I$;
         Knowledge Graph $\mathcal{G}$;
         Toolset $\mathcal{T}$;
         Memory $\mathcal{M}$;
         Schema information $S$

\ENSURE Final Verdict $y$;
        Evidence Chain $\mathcal{E}$

\STATE
$\mathcal{C}_{\mathrm{mem}}
\gets
\text{\textsc{RetrieveExpertDemos}}(\tau,\mathcal{M})$

\STATE
$\mathcal{P}_{\mathrm{init}}
\gets
\text{\textsc{GenerateInitialPlan}}
(\tau_q, \mathcal{S}, \mathcal{C}_{mem}, \mathcal{T}) \big)$

\STATE $h_0 \gets \emptyset$

\STATE
$\mathcal{H}_1
\gets
(I,\mathcal{C}_{\mathrm{mem}},
\mathcal{P}_{\mathrm{init}},h_0)$

\STATE $t \gets 1$

\WHILE{$t \leq T_{\max}$}

    \STATE
    $th_t
    \sim
    \pi_\theta
    \bigl(
        \text{Think}
        \mid
        \mathcal{H}_t
    \bigr)$

    \STATE
    $ac_t
    \gets
    (a_t,p_t)
    \sim
    \pi_\theta
    \bigl(
        \text{Act}
        \mid
        \mathcal{H}_t,th_t
    \bigr)$

    \IF{$a_t$ is \textsc{Finish[Answer]}}
        \RETURN
        $(y,\mathcal{E}) \gets \text{\textsc{ParseVerdict}}(p_t)$
    \ENDIF

    \STATE
    $obs_t
    \gets
    \text{\textsc{Env}}
    (a_t,p_t,\mathcal{G},\text{Web})$

    \STATE
    $h_t
    \gets
    h_{t-1}
    \cup
    \{(th_t,ac_t,obs_t)\}$

    \STATE
    $\mathcal{H}_{t+1}
    \gets
    (I,\mathcal{C}_{\mathrm{mem}},
    \mathcal{P}_{\mathrm{init}},h_t)$

    \STATE $t \gets t+1$

\ENDWHILE

\RETURN
$\displaystyle
\arg\max_{y,\mathcal{E}}
P_{\pi_\theta}
\left(
    y,\mathcal{E}
    \mid
    \mathcal{H}_{T_{\max+1}}
\right)$

\end{algorithmic}
\end{algorithm}

\subsection{Hybrid Knowledge Toolset}
\label{ssec:Hybrid Knowledge Toolset}

To bridge the gap between KG structural information and external textual semantics, we design a hybrid knowledge toolset $\mathcal{T}$ with five atomic capabilities, grouped into two categories.

\subsubsection{KG Internal Structure Tools}

These tools exploit the inherent logical constraints and topological structures of KGs.

\begin{itemize}[leftmargin=1em, labelsep=0.5em, itemindent=0pt, parsep=0.5ex]

\item \textbf{KG Definition Tool:}
Given an entity $e$ or relation $r$, this tool queries the underlying metadata. For entities, it returns labels, descriptions, and aliases; for relations, it returns semantics, domain, and range. It provides explicit definitions for entities and relations during verification.

\item \textbf{KG Neighbor Tool:}
Given an entity $e$ and a target relation $r$, this tool retrieves the top-$k$ most relevant entity $e$'s triples under relation $r$ direction using dense retrieval based on cosine similarity. The retrieved neighbors help the agent understand the contextual semantics of entities along a specific relational direction.

\item \textbf{KG Path Tool:}
Given a head entity $h$ and a tail entity $t$, this tool applies bi-directional BFS to retrieve relational paths connecting them within $n \leq 3$ hops. Multi-hop paths provide structured evidence for implicit reasoning. Formally:
\begin{equation*}
\begin{split}
\text{Paths}(h,t) &= \{(h, r_1, x_1),\ldots,(x_{n-1}, r_n, t) \\
&\quad \mid \exists x_1,\ldots,x_n \in \text{KG},\ n \leq 3\}
\end{split}
\end{equation*}

\end{itemize}

\subsubsection{External Semantic Tools}

These tools complement KG sparsity with open-world textual evidence.

\begin{itemize}[leftmargin=1em, labelsep=0.5em, itemindent=0pt, parsep=0.5ex]

\item \textbf{Wiki Evidence Tool:}
This tool directly accesses Wikipedia and supports two modes. (1) Entity Mode: retrieves summaries and key attributes for a given entity, providing encyclopedic background
knowledge. (2) Co-occurrence Mode: given an entity pair $(h,t)$, retrieves sentence-level contexts where the two entities co-occur within a word-distance threshold $\tau$, providing explicit textual evidence for hidden relations.

\item \textbf{Web Evidence Tool:}
This tool calls commercial search engine APIs (e.g., Google/Bing), takes an agent-generated natural language query as input, and returns top-$k$ web snippets. It is particularly useful for verifying long-tail knowledge, emerging entities, and time-sensitive facts.

\end{itemize}

\subsubsection{Hybrid Multi-stage Retrieval}
Obviously, retrieval plays an important role in the above tool design.
We implement a hybrid multi-level retrieval mechanism for all retrieval-based tools.
This mechanism combines keyword-based exact matching with vector-based semantic re-ranking. Exact matching preserves high-precision lexical hits, while semantic re-ranking recovers alias and paraphrase matches that lexical retrieval may miss.

\section{Experiments and Results}

\begin{table*}[t]
\centering

\begingroup
\footnotesize
\setlength{\tabcolsep}{2.2pt}
\renewcommand{\arraystretch}{1.10}

\begin{tabularx}{\textwidth}{
  @{}
  >{\raggedright\arraybackslash}p{0.325\textwidth}
  @{\hspace{1pt}}
  *{8}{>{\centering\arraybackslash}X}
  @{}
}

\toprule
\multirow{2}{*}{\textbf{Method}}
& \multicolumn{4}{c}{\textbf{FB15K-237}}
& \multicolumn{4}{c}{\textbf{Wikidata5M-Ind}} \\
\cmidrule(lr){2-5}
\cmidrule(lr){6-9}
& \textbf{Acc.}
& \textbf{F1}
& \textbf{Prec.}
& \textbf{Rec.}
& \textbf{Acc.}
& \textbf{F1}
& \textbf{Prec.}
& \textbf{Rec.} \\
\midrule

\multicolumn{9}{@{}l}{\textit{Graph Embedding-based}}\\[-0.4ex]

\meth{\mbox{TransE~\citep{Intro_TransE}}}
& 66.4
& 73.7
& 60.5
& 94.2
& 49.6
& 49.2
& 49.6
& 48.8 \\

\meth{\mbox{DistMult~\citep{Re_DistMult}}}
& 61.2
& 70.9
& 56.7
& \gb{94.4}
& 50.2
& 50.7
& 51.4
& 50.1 \\

\meth{\mbox{RotatE~\citep{Intro_RotatE}}}
& 66.6
& 74.4
& 60.3
& \best{97.0}
& 50.8
& 51.3
& 50.5
& 52.2 \\

\midrule
\multicolumn{9}{@{}l}{\textit{PLM-based}}\\[-0.4ex]

\meth{\mbox{KG-BERT~\citep{Intro_KG_BERT}}}
& 66.3
& 72.0
& 62.2
& 85.5
& --
& --
& --
& -- \\

\meth{\mbox{SimKGC~\citep{Re_SimKGC}}}
& 69.8
& 72.5
& 66.6
& 79.4
& 78.4
& \gb{78.9}
& 76.9
& \gb{81.1} \\

\midrule
\multicolumn{9}{@{}l}{\textit{LLM zero-shot}}\\[-0.4ex]

\meth{\mbox{GPT-3.5-turbo~\citep{Exper_InstructGPT}}}
& 69.7
& 73.4
& 65.4
& 83.4
& 73.8
& 72.6
& 76.1
& 69.4 \\

\meth{\mbox{GPT-4o~\citep{Exper_gpt4}}}
& 77.3
& 73.9
& \gb{86.8}
& 64.4
& 77.6
& 72.6
& \gb{93.1}
& 59.6 \\

\meth{\mbox{Qwen3-max~\citep{Exper_Qwen3}}}
& 76.6
& 73.7
& 84.3
& 65.4
& 74.5
& 67.7
& 92.4
& 53.4 \\

\midrule
\multicolumn{9}{@{}l}{\textit{LLM inference-augmented}}\\[-0.4ex]

\meth{\mbox{CoT~\citep{Exper_CoT}}}
& 81.4
& 80.1
& 86.0
& 75.0
& 80.5
& 77.4
& 92.0
& 66.8 \\

\meth{\mbox{Self-Consistency~\citep{Exper_CS}}}
& 80.8
& 79.5
& 85.2
& 74.6
& \gb{80.8}
& 77.9
& 91.4

& 68.0 \\

\meth{\mbox{KGValidator~\citep{Re_KGValidator}}}
& \gb{83.0}
& \gb{81.0}
& --
& --
& --
& --
& --
& -- \\

\midrule
\multicolumn{9}{@{}l}{\textit{Our method}}\\[-0.4ex]

\meth{\textbf{SHARP}}
& \best{\meanstd{87.1}{0.9}}
& \best{\meanstd{86.5}{1.0}}
& \best{\meanstd{90.8}{0.9}}
& \meanstd{82.7}{1.0}
& \best{\meanstd{93.6}{0.4}}
& \best{\meanstd{93.3}{0.4}}
& \best{\meanstd{97.9}{0.3}}
& \best{\meanstd{89.1}{1.0}} \\

\bottomrule
\end{tabularx}

\endgroup

\caption{
Results on FB15K-237 and Wikidata5M-Ind. All methods are evaluated on the same three independently sampled test sets; SHARP values are reported as mean ± standard deviation. Cross-paradigm methods retain their native evidence access “–” denotes unavailable results. Bold indicates the best-performing model; underlining indicates the second-best model.
}
\label{Tab:main}
\end{table*}

\subsection{Experiment Setting}

\paragraph{Benchmarks and Negative Sampling}
We evaluate SHARP on two benchmarks with distinct properties. FB15K-237~\cite{Ex_FB}, derived from Freebase, removes inverse relations and emphasizes reasoning-oriented link prediction with multi-hop inference. Wikidata5M-Inductive~\cite{Re_KEPLER}, derived from Wikidata, is a recent inductive benchmark where training and test entities are disjoint. It features extreme sparsity and a pronounced long-tail distribution. 

For the main experiments, we independently sampled each benchmark three times, resulting in three test sets of 1,000 instances each. Because the original benchmarks contain only positive triples, we converted 500 positive triples in each test set into negative ones using type-constrained negative sampling. Given a positive triple $\tau = (h, r, t)$, we replace either the head or tail entity with another entity of the same type to obtain a negative triple $\tau^-$. This strategy avoids type-incompatible negatives, such as ("Obama," "born in," "banana"), and yields a nontrivial but controlled evaluation setting. To ensure the validity of negative samples, we manually verified the generated negative triples and removed cases that could still be supported by factual evidence, ensuring that they represent incorrect statements. Unless otherwise specified, all subsequent results are averaged over the three test sets. The resulting balanced sets evaluate discrimination between verified positives and type-compatible constructed negatives; they are not intended to reproduce the natural prevalence or full error distribution of deployed KGs.

\paragraph{Baselines}
We compare SHARP with representative baselines from three mainstream paradigms: 

\begin{itemize}[leftmargin=1em, labelsep=0.5em, itemindent=0pt, parsep=0.5ex]

\item \textbf{Graph Embedding methods}: TransE~\cite{Intro_TransE}, DistMult~\cite{Re_DistMult}, and RotatE~\cite{Intro_RotatE}; 

\item \textbf{PLM-based methods:} KG-BERT~\cite{Intro_KG_BERT} and SimKGC~\cite{Re_SimKGC}; 

\item \textbf{LLM-based methods:} (1) \textbf{Zero-shot Inference:} GPT-3.5-Turbo~\cite{Exper_InstructGPT}, GPT-4o~\cite{Exper_gpt4}, and Qwen3-Max~\cite{Exper_Qwen3}; (2) \textbf{Inference-Augmented Frameworks:} Chain-of-Thought (CoT)~\cite{Exper_CoT}, Self-Consistency (SC)~\cite{Exper_CS}, and the KG-validation-specific KGValidator ~\cite{Re_KGValidator}.

\end{itemize}

Baseline configurations are in the Appendix.

\paragraph{Evaluation metrics}
We use Accuracy and F1-score as the main evaluation metrics, with precision and recall as supplementary metrics. Accuracy reflects overall correctness on the balanced test set, while F1 provides a more robust measure by jointly considering precision and recall.

\paragraph{Leak prevention constraint}
To mitigate information leakage, we remove the target triple, inverse relations, and alias-equivalent statements from the outputs of the KG Neighbor Tool and KG Path Tool before they are provided to the agent. This constraint does not exclude knowledge encoded during pretraining or retrieved from public webpages, as such information constitutes a standard evidence source for retrieval-based knowledge verification rather than task-specific leakage.

Further experimental setting details such as
model hyperparameters, vector retrieval configurations are provided in the Appendix.

\begin{figure}[tbp]
    \centering
    \includegraphics[width=\linewidth]{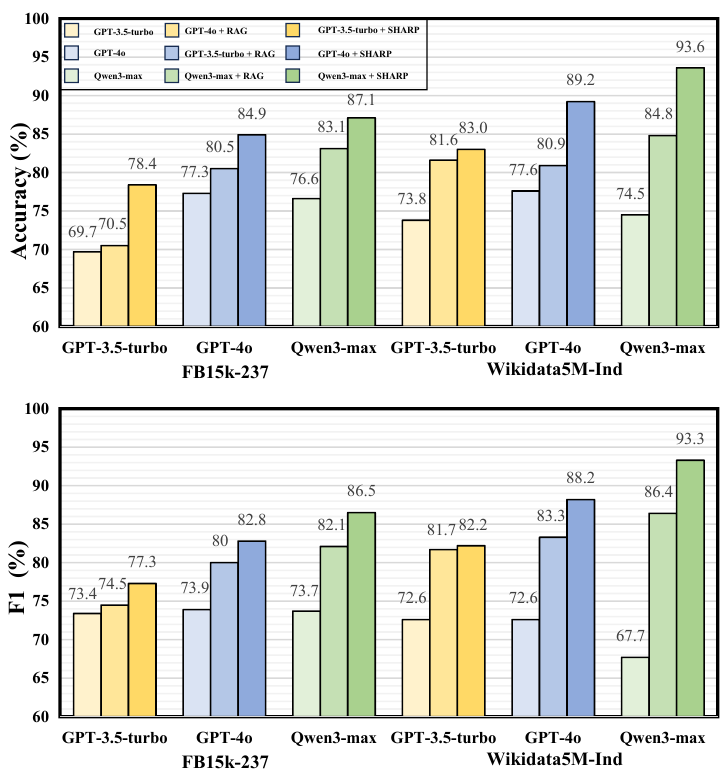}    
    \caption{Comparison of LLM vs. LLM+RAG vs. LLM+SHARP on FB15K-237 and Wikidata5M-Ind. RAG uses the same evidence sources as SHARP but performs one non-interactive verification pass.}    
    \label{FIG:vs}
\end{figure}

\subsection{Main Result: SHARP vs. Other Paradigms}
Table~\ref{Tab:main} compares Qwen3-Max-based SHARP with representative paradigms on two benchmarks. Because these cross-paradigm methods differ in evidence access and computation, these are contextual rather than budget-matched comparisons; the same-source RAG comparison in Figure~\ref{FIG:vs} provides key evidence for SHARP’s orchestration advantage. SHARP achieves the best accuracy and F1 on both benchmarks, with especially large gains on inductive Wikidata5M-Ind.

\paragraph{\textbf{Compared with graph embedding models,}}  SHARP brings substantial and consistent improvements, especially in the inductive scenario where entities and triples are unseen during training. Compared with the strongest embedding baseline, SHARP shows absolute Accuracy/F1 gains of 20.5/12.1 percentage points over RotatE on FB15K-237 and 42.8/42.0 points on Wikidata5M-Ind. This is because SHARP integrates external tools and evidence signals, overcoming the limitations of models relying solely on triple structural information.

\paragraph{\textbf{Compared with PLM-based methods,}} SHARP also achieves consistently better performance. Relative to SimKGC, the absolute Accuracy/F1 gains are 17.3/14.0 percentage points on FB15K-237 and 15.2/14.4 points on Wikidata5M-Ind, respectively. Instead of relying on static triple text, our framework actively retrieves information and reasons via the agent architecture to form effective multi-hop reasoning chains.

\paragraph{\textbf{Compared with zero-shot and inference-augmented LLM methods,}} SHARP exceeds KGValidator on FB15K-237 by 4.1 Accuracy and 5.5 F1 points and substantially outperforms all LLM baselines on Wikidata5M-Ind. These results suggest that relying only on LLM parametric knowledge or lightweight reasoning augmentation is prone to hallucination and noisy inference, whereas SHARP mitigates such noise through graph schema-aware dynamic retrieval and reasoning.



\subsection{Agentic Reasoning vs. Standard RAG}

SHARP incorporates richer internal and external evidence than conventional methods. To address potential concerns regarding unequal information access or data leakage, we construct a controlled Standard RAG baseline that uses exactly the same data sources and retrieval modules as SHARP. Figure~\ref{FIG:vs} compares Zero-shot LLM, Standard RAG, and SHARP across two benchmarks and three backbones: RAG consistently improves on zero-shot inference, while SHARP further improves Accuracy and F1 in every setting. Equalizing token budgets would either truncate SHARP's interaction or turn RAG into a multi-step agent; we therefore match evidence sources, not interaction budgets, to test whether source access alone suffices.

These results confirm the effectiveness of the retrieval component in our Hybrid Knowledge Toolset and, more importantly, show that SHARP's improvements cannot be attributed merely to additional data. Instead, its gains primarily arise from iterative multi-step reasoning. Unlike standard RAG, which may struggle to identify critical evidence and perform deep reasoning in knowledge-intensive contexts, SHARP employs dynamic planning and stepwise verification to distill relevant information and improve verification accuracy.

\begin{table}[tbp]
\centering

\resizebox{\linewidth}{!}{
\begin{tabular}{lcccc}
\toprule
\multirow{2}{*}{\textbf{Method}} & \multicolumn{2}{c}{\textbf{FB15K-237}} & \multicolumn{2}{c}{\textbf{WD5M-Ind}} \\
\cmidrule(lr){2-3} \cmidrule(lr){4-5}
 & \textbf{Acc} & \textbf{F1} & \textbf{Acc} & \textbf{F1} \\
\midrule
\textbf{SHARP} & \best{87.1} & \best{86.5} & \best{93.6} & \best{93.3} \\
\midrule
w/o Memory Augmentation &81.0 & 80.9  & 86.8  &86.8  \\
w/o Schema-Aware Planning & 77.9  & 76.7  & 79.1  & 76.7  \\

w/o KG Tools          & 78.4 & 77.2  & 83.7  & 81.9  \\
w/o External Tools    & 75.7  & 70.6  & 80.5  & 76.1 \\
\bottomrule
\end{tabular}
}
\caption{Component ablation results on benchmarks}
\label{Tab:ablation}
\end{table}

\subsection{Ablation Study}
Table~\ref{Tab:ablation} removes each core component in turn. Every removal lowers performance on both benchmarks, showing dependence on all four components in the evaluated configuration:


\begin{itemize}[leftmargin=1em, labelsep=0.5em, itemindent=0pt, parsep=0.5ex]
    \item \textbf{w/o Memory Augmentation}: plans from the model's triple understanding without retrieved memory.
    \item \textbf{w/o Schema-Aware Planning}: uses expert memory traces directly for iterative reasoning.
    \item \textbf{w/o KG Tools}: removes definition, neighborhood, and path tools for structured knowledge.
    \item \textbf{w/o External Tools}: removes Wikipedia and web evidence retrieval.
\end{itemize}

\paragraph{Analysis of w/o Memory Augmentation}
Removing memory augmentation leads to approximately 6 percentage points drops in both accuracy and F1 on the benchmarks. This indicates that memory augmentation allows the agent to leverage past reasoning experiences and optimal inference paths, guiding both initial planning and multi-step verification.

\paragraph{Analysis of w/o Schema-Aware Planning}
Removing schema-aware planning reduces Accuracy by 9.2 points on FB15K-237 and 14.5 points on Wikidata5M-Ind, and F1 by 9.8 and 16.6 points. The agent struggles to trace reasoning chains within complex entity networks, often producing redundant loops or being misled by irrelevant information.

\paragraph{Analysis of w/o KG Tools}
Removing KG tools reduces Accuracy by 8.7 and 9.9 percentage points and F1 by 9.3 and 11.4 points on FB15K-237 and Wikidata5M-Ind, respectively. This confirms that graph-structured neighborhood and path information provide critical contextual constraints, mitigating hallucinations during fact verification.

\paragraph{Analysis of w/o External Tools}
Removing external tools reduces Accuracy by 11.4 and 13.1 percentage points and F1 by 15.9 and 17.2 points on FB15K-237 and Wikidata5M-Ind, respectively. External knowledge retrieval provides large-scale, current evidence beyond the triple, underscoring the value of integrating internal and external knowledge.

\begin{figure}[t]
    \centering
    \includegraphics[width=\linewidth]{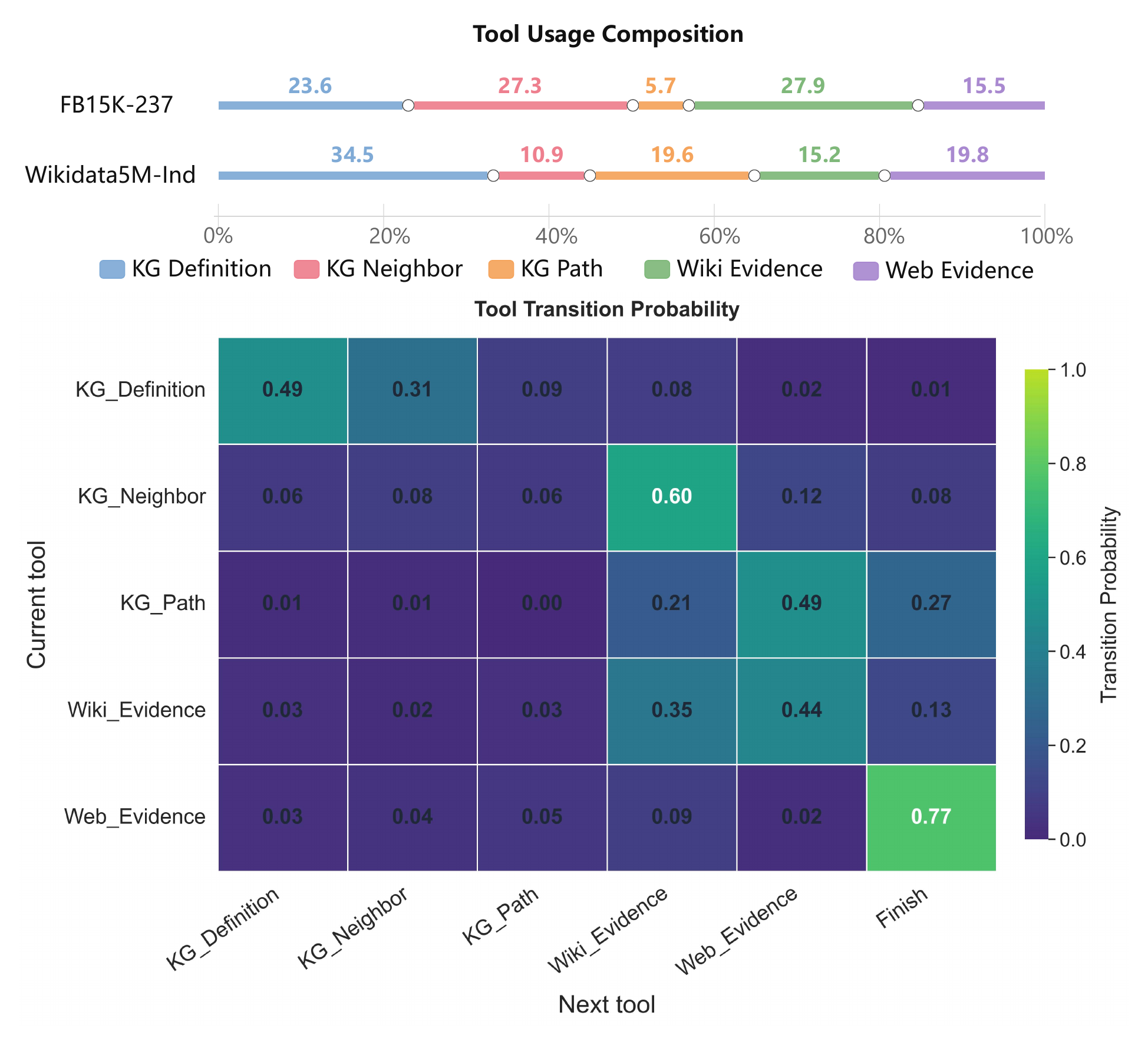}    
    \caption{Tool-use behavior analysis. The upper panel presents the tool invocation distribution on FB15K-237 and Wikidata5M-Ind, while the lower panel shows transition probabilities between consecutive tool calls. }    
    \label{FIG:tool_usage}
\end{figure}

\subsection{Tool-use Behavior Analysis}
Figure~\ref{FIG:tool_usage} reports both the tool invocation distribution and the transition probabilities between consecutive tool calls. The usage distribution shows clear dataset-specific behavior. On FB15K-237, SHARP relies most on Wiki Evidence (27.9\%) and KG Neighbor (27.3\%), indicating that verification often requires cross-checking local graph context with external textual evidence. In contrast, on Wikidata5M-Ind, KG Definition becomes dominant (34.5\%), while KG Path (19.6\%) and Web Evidence (19.8\%) are used more often, reflecting the sparse, long-tail inductive setting.

The transition matrix further reveals a staged verification pattern. Definition calls are commonly followed by another definition query (0.49) or a neighbor query (0.31), establishing basic entity and relation semantics before deeper verification. Structural tools then tend to trigger external validation, especially Neighbor $\rightarrow$ Wiki (0.60) and Path $\rightarrow$ Web (0.49). Finally, Web Evidence most often leads to termination (0.77), suggesting that open-world evidence serves as the final confirmation step. Overall, SHARP performs adaptive tool orchestration rather than uniform retrieval.

\begin{table}[t]
\centering
\resizebox{\linewidth}{!}{
\begin{tabular}{lcccc}
\toprule
Split & $n$ & Relevant & Sufficient & Support \\
\midrule
Wikidata5M-Ind & 50 & 100\% & 86\% & 84\% \\
FB15K-237 & 50 & 98\% & 84\% & 84\% \\
Overall & 100 & 99\% & 85\% & 84\% \\
\bottomrule
\end{tabular}
}
\caption{Evidence-chain audit of 100 correctly predicted triples.}
\label{tab:evidence-consistency}
\end{table}

\subsection{Evidence-Chain Quality Assessment}
We further evaluate the quality of SHARP's generated evidence chains through manual assessment. We randomly sampled 50 correctly predicted triples from each benchmark (100 total) and evaluated whether the generated evidence was relevant to the target triple, sufficient to justify the reported verdict, and supportive of that verdict. Conditional on a correct prediction, the traces achieve 99\% relevance, 85\% sufficiency, and 84\% verdict support. 
\section{Discussion}

\paragraph{Why SHARP helps.}
Controlled RAG comparisons, component removals, and tool-use patterns are jointly consistent with a benefit from adaptive acquisition and combination of structural and textual evidence beyond source access alone. The recurring structure-to-text transitions further indicate that SHARP follows a staged verification process rather than uniform retrieval. Nevertheless, these results should be viewed as supporting, rather than causally isolating, the effect of planning, since SHARP also employs sequential tool calls and an adaptive computation budget.
\paragraph{Where verification remains difficult.}
Despite manual checks that remove supported negatives, historical benchmark labels can conflict with current Wikidata or web evidence. The Appendix shows more retrieval and relation-granularity errors on FB15K-237, versus more entity-grounding and temporal/contextual mismatches on Wikidata5M-Ind. Accuracy therefore measures agreement with constructed labels under our retrieval protocol, not timeless factual correctness.
\paragraph{What inspectability enables.}
SHARP’s evidence chains expose executed queries, returned observations, and supporting evidence, but do not guarantee source correctness or causal faithfulness.  Given the additional tool interactions and costs reported in the Appendix, SHARP is currently better suited to selective or human-supervised auditing; conflicting or insufficient evidence should trigger abstention and curator review rather than automatic KG updates.

\section{Conclusion}


In this work, we introduced SHARP, a training-free framework casting KG triple verification as dynamic plan--retrieve--reason over structural and open-world evidence. By combining schema-aware planning, memory augmentation, and a hybrid knowledge toolset, SHARP stabilizes reasoning and effectively integrates graph-structural logic with external semantic evidence. Across three independently sampled test sets on FB15K-237 and Wikidata5M-Ind, SHARP consistently outperforms strong  baselines. Controlled RAG comparisons and ablations further suggest that its gains reflect planning, memory, and iterative tool use beyond retrieval access. SHARP also produces traceable, fact-grounded reasoning chains, demonstrating effectiveness and interpretability for complex triple verification.

\bibliography{aaai2027}

\begin{center}
{\LARGE\bfseries Appendix}
\end{center}

\section{Experimental Settings}
\label{sec:ex_setting}
\subsection{Core Agent Configuration}
We employ Qwen3-Max as the backbone controller for SHARP, accessed through an OpenAI-compatible Python SDK interface. We set the decoding temperature to 0 to reduce sampling variability. The stop sequence is defined as ``[Observation]'' to enforce the ReAct interaction format, and the maximum number of interaction turns is $T_{\max}=10$. Temperature zero does not make the pipeline fully deterministic: model-service updates, concurrent execution, and live KG or web retrieval can still change a returned trace. Accordingly, the random seeds below govern evaluation-set construction rather than guaranteeing bitwise-identical API responses.

\subsection{Vector Retrieval and Indexing}
For all vector-based retrieval tasks—including trajectory retrieval from the Memory Bank and semantic re-ranking of web snippets—we utilize the all-MiniLM-L6-v2 model for efficient text encoding. We employ the FAISS library to construct dense vector indexes, enabling high-speed similarity search.  Retrieval uses cosine similarity; memory retrieval returns $k=3$ trajectories.

\begin{table}[htbp]
\centering
\small
\setlength{\tabcolsep}{4pt}
\renewcommand{\arraystretch}{1.05}
\begin{tabular}{@{}lcccc@{}}
\toprule
\textbf{Benchmark} & \textbf{Seeds} & \textbf{Size/seed} & \textbf{Positive} & \textbf{Negative} \\
\midrule
FB15K-237 & 3 & 1,000 & 500 & 500 \\
Wikidata5M-Ind & 3 & 1,000 & 500 & 500 \\
\bottomrule
\end{tabular}
\caption{Evaluation-set construction. Counts are per independently sampled seed.}
\label{tab:evaluation-sets}
\end{table}


\subsection{Evaluation Set Construction and Repeated Evaluation}
\label{sec:data-construction}

As shown in Table~\ref{tab:evaluation-sets},for each benchmark, we construct three independently sampled evaluation sets using seeds 2024, 2025, and 2026. Each set contains 1,000 balanced instances: 500 positive triples and 500 type-constrained negatives. We generate each negative by replacing either the head or the tail of a positive triple with an entity of the same type, and manually remove generated negatives that remain factually supportable. Unless stated otherwise, we apply the same evaluation protocol to all three sets and report each SHARP aggregate as the mean $\pm$ sample standard deviation across them.

The seeds control evaluation-set sampling and negative construction, but do not freeze remote model or retrieval services. Accordingly, the three results represent independently sampled evaluation replicates rather than deterministic reruns on a fixed test set. The reported Accuracy and F1 therefore measure discrimination between verified positives and constructed, type-compatible negatives; they should not be interpreted as estimates of the natural error prevalence in a deployed KG.

\subsection{Knowledge Environment and Tool Implementation}
\paragraph{Knowledge Retrieval Environment.}
We use Wikidata as a unified KG interface for both benchmarks, as entities in their source knowledge graphs can be aligned with Wikidata. Specifically, Wikidata5M is directly derived from Wikidata, whereas FB15K-237 is derived from Freebase, a knowledge base that predates Wikidata and for which mappings from Freebase entity identifiers to Wikidata identifiers are available. We apply the disambiguation procedure described below to resolve missing or potentially ambiguous mappings. The resolved Wikidata identifiers are then used to construct SPARQL queries for retrieving structural evidence.

\paragraph{Entity/Relation Disambiguation.} To map natural language terms to Wikidata IDs (PID/QID), we adopt a two-stage strategy: (1) \textbf{Keyword Matching:} We first attempt to retrieve IDs via exact keyword search; (2) \textbf{Semantic Matching:} If keyword retrieval fails, we fall back to vector-based matching using the aforementioned encoder to identify the ID with the closest semantic meaning.

\paragraph{Anti-Leakage Mechanism.} During KG Neighbor and KG Path retrieval, we remove (i) the exact target triple, (ii) its inverse-direction equivalent, and (iii) alias-equivalent statements that resolve to the same entity pair and relation. This filtering applies only to tool outputs before they enter the agent context. It does not remove facts encoded in the backbone model or retrieved from public web pages; we therefore interpret the experiment as retrieval-based verification under an explicit information boundary, not as a closed-book leakage test.

\paragraph{Tool Return Limits.} To balance information density and context window usage, the KG tools (Neighbor and Path) return the Top-20 most relevant triples or paths, while the Web Evidence Tool (utilizing the Google Search API) returns the Top-5 text snippets.



\begin{table*}[t]
\centering
\small 

\renewcommand{\arraystretch}{1.1} 
\begin{tabularx}{\textwidth}{@{} 
    >{\raggedright\arraybackslash}p{2.7cm} 
    >{\raggedright\arraybackslash}p{4.0cm} 
    X 
    @{}}
\toprule
\textbf{Tool API Name} & \textbf{Parameters} & \textbf{Output Evidence \& Core Usage} \\
\midrule

\multicolumn{3}{@{}l}{\textbf{1. Internal Structural Tools (KG)}} \\
\texttt{KG\_Definition} & 
\texttt{entity} \textit{or} \texttt{relation}: str & 
Returns schema metadata (labels, domains, ranges) to establish basic definitions for nodes/edges. \\
\addlinespace 
\texttt{KG\_Neighbor} & 
\texttt{entity}, \texttt{relation}: str & 
Returns top-20 semantically correlated target entities to analyze local subgraphs and co-occurrences. \\
\addlinespace
\texttt{KG\_Path} & 
\texttt{entity\_a}, \texttt{entity\_b}: str & 
Returns explicit 1 to 3-hop relational paths to verify structural connectivity and implicit links. \\

\midrule
\multicolumn{3}{@{}l}{\textbf{2. External Semantic Tools}} \\

\texttt{Wiki\_Evidence} & 
(1) \texttt{entity}: str \newline 
(2) \texttt{entity\_a}, \texttt{entity\_b}: str & 
(1) Returns Wikipedia summaries/attributes to provide encyclopedic background for a single entity. \newline
(2) Returns sentences mentioning both entities to discover explicit textual links. \\
\addlinespace
\texttt{Web\_Evidence} & 
\texttt{question}: str & 
Returns top-5 Google search snippets as a final fallback validation for long-tail or emerging facts. \\

\bottomrule
\end{tabularx}
\caption{Overview of the Hybrid Knowledge Toolset API definitions.}
\label{tab:toolset}
\end{table*}

\subsection{Experimental Setup of Baselines}
\label{sec:baseline}
Baselines use official or standard implementations: KGE models are implemented with OpenKE, PLM methods use their official codebases, and LLM approaches use their official APIs. For trainable baselines, we retain the paper-recommended configurations; for API-based LLM baselines, we use the provider defaults unless otherwise specified.  These cross-paradigm comparisons retain their native evidence access and computation and are therefore contextual rather than budget-matched. The controlled Standard-RAG comparison in the main paper uses the same evidence sources as SHARP and isolates the benefit of iterative orchestration more directly.

Given the significant disparity in the output paradigms across various baselines, we implement a rigorous alignment evaluation protocol to ensure a fair comparison: (1) For KGE baselines outputting continuous scores (e.g., TransE), we search for the optimal threshold $\delta$ on the validation set to binarize these scores into final Boolean predictions. (2) For generative approaches based on LLMs and Agents, we directly parse their explicitly generated textual responses. Notably, to maintain strict evaluation standards, any refusal to answer or formatting violation is strictly penalized as an incorrect prediction.


\subsection{Evaluation metrics}
\label{sec:eva}

We adopt four fundamental metrics widely adopted in classification tasks to quantify model performance: Accuracy, Precision, Recall, and F1-score. Their formal definitions and mathematical formulations are as follows, where: $TP$ (true positives) denotes the count of positive instances correctly predicted; $TN$ (true negatives) denotes the count of negative instances correctly predicted; $FP$ (false positives) denotes the count of negative instances incorrectly predicted as positive; and $FN$ (false negatives) denotes the count of positive instances incorrectly predicted as negative. 
Accuracy measures the overall proportion of correct predictions among all instances: $$\text{Accuracy} = \frac{TP + TN}{TP + TN + FP + FN}$$ 
Precision quantifies the reliability of positive predictions, i.e., the proportion of predicted positives that are actually positive: $$\text{Precision} = \frac{TP}{TP + FP}$$ 
Recall quantifies the completeness of positive predictions, i.e., the proportion of actual positives that are correctly identified: $$\text{Recall} = \frac{TP}{TP + FN}$$ 
F1-score is the harmonic mean of Precision and Recall, balancing their trade-off to provide a single metric for overall performance: $$\text{F1-score} = 2 \times \frac{\text{Precision} \times \text{Recall}}{\text{Precision} + \text{Recall}}$$

\begin{figure}[h]
    \centering
    \includegraphics[width=\linewidth]{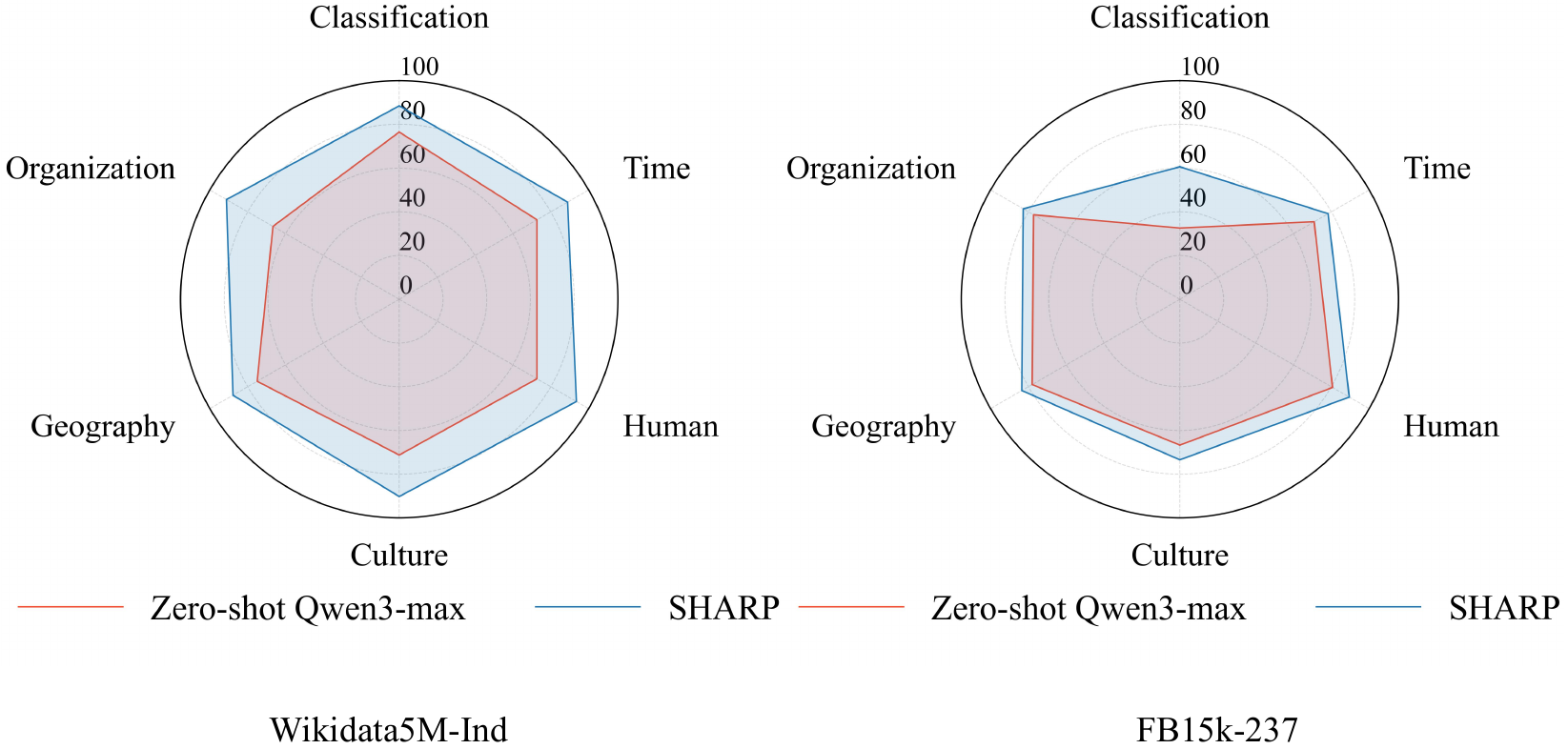}
    \caption{F1 comparison between SHARP and zero-shot Qwen3-Max across six schema categories.}
    \label{Fig:schema_f1}
\end{figure}

\section{Toolset API Interface}
\label{sec:tool-api}
Table~\ref{tab:toolset} summarizes the API interface of the Hybrid Knowledge Toolset used in SHARP. These tools are explicitly defined as API functions that the LLM agent can directly invoke through executable JSON-formatted calls. For each API, the table specifies its callable name, required input parameters, and returned evidence. The toolset includes internal KG tools for retrieving schema definitions, local neighbors, and multi-hop relational paths, as well as external semantic tools for obtaining Wikipedia evidence and web search snippets. This unified interface clarifies both the format and the effect of tool invocation in the ReAct reasoning loop, where each returned observation supports structural-semantic cross-verification for knowledge graph triple verification.

\begin{figure*}[t]
\centering
\begin{tcolorbox}[colback=gray!5!white,colframe=gray!70!black,halign title=center, center title,title=\textbf{Prompt 1: Schema-Aware Planning}, width=\textwidth]
\small 
\textbf{System Instruction:} \\
You are a specialized Knowledge Graph Verification Strategist.
Your goal is to provide a Verification Plan for a given triple (Head, Relation, Tail) by analyzing the pattern characteristics of the triple to be tested.

\textbf{[Tool Definitions]} \\
(Note: The full tool definitions can be found in Appendix~\ref{sec:tool-api}. They are omitted here for the sake of brevity.)

\textbf{Refer to the verification case:} \\
This is a complete verification process for a task similar to the task to be verified, which can be used as a reference when formulating the initial verification plan. \\
{Memory Trajectory}

\textbf{Output Format} \\
Please strictly follow this format: \\
=== Strategic Plan === \\
{[Step 1 description using ToolName and Toolparam]} \\
{[Step 2 description using ToolName and Toolparam]} \\
...

\textbf{User Input:} \\
Target Triple: \{triple\}
\end{tcolorbox}
\caption{The prompt template for the Schema-Aware Planning phase.}
\label{fig:prompt_plan}
\end{figure*}

\section{Schema-wise Performance Analysis}

To further examine whether SHARP effectively supports triple verification under diverse relation schemas, we group test triples into six graph-schema categories according to the memory-bank taxonomy described  in the Schema-Aware Initialization section.
Figure~\ref{Fig:schema_f1} reports F1 computed separately within each category. SHARP outperforms the zero-shot LLM baseline in all six descriptive slices, indicating that the observed gain is not confined to one predefined schema category. Because category sizes and difficulties may differ, this figure is a heterogeneity check rather than a statistical test of equal gains across categories.

\section{Prompt Templates}
\label{sec:prompt}

We provide role-level prompt templates to document the roles, interaction protocol, and output format. Braced fields denote run-time values and ellipses mark descriptive text omitted from the figure for space. The exact machine-readable prompt strings, tool schemas, parsing rules, and stop conditions should be included in the accompanying Code and Data Supplement; the figures below are not intended to substitute for those files.
The \textbf{Schema-Aware Planning Prompt} (Figure~\ref{fig:prompt_plan}) is used in Phase 1 to generate the initial verification strategy.
The \textbf{Reasoning Instruction Prompt} (Figure~\ref{fig:prompt_reasoning}) serves as the system instruction for the core ReAct loop in Phase 2, defining tool signatures and interaction protocols.
Finally, the \textbf{Mandatory Judgment Prompt} (Figure~\ref{fig:prompt_judgment}) is triggered in Phase 3 if the agent exceeds the maximum step limit, forcing a probabilistic verdict.

\begin{table}[tbp]
  \centering

  \small
  \setlength{\tabcolsep}{3.5pt}
  \renewcommand{\arraystretch}{1.08}
  \begin{tabular}{lcccc}
    \toprule
    \multirow{2}{*}{\textbf{Dataset}}
    & \multicolumn{4}{c}{\textbf{Avg.}} \\
    \cmidrule(lr){2-5}
    & \multicolumn{1}{c}{\textbf{Amort. Time}}
    & \multicolumn{1}{c}{\textbf{In Tokens}}
    & \multicolumn{1}{c}{\textbf{Out Tokens}}
    & \multicolumn{1}{c}{\textbf{Cost}} \\
    \midrule
    FB15K-237      & 1.86 & 23,031.39 & 2,275.24 & \$0.011 \\
    Wikidata5M-Ind & 1.26 & 13,099.55 & 1,254.77 & \$0.006 \\
    \bottomrule

  \end{tabular}
      \caption{Amortized wall-clock time, token use, and API cost per completed query under 50-worker batch execution.}
  \label{Tab:cost}
\end{table}

\section{Efficiency and Cost Analysis}

To evaluate and optimize operational efficiency, we implemented a multi-threading concurrency mechanism configured with 50 threads. We use 50 concurrent worker threads for the reported measurements. As presented in Table~\ref{Tab:cost}, the average inference latency of SHARP on FB15K-237 and Wikidata5M-Ind is 1.86s and 1.26s, respectively. 

Regarding API costs and token consumption, statistics indicate that the average number of agent interactions per query is 9.8 on FB15K-237 and 6.6 on Wikidata5M-Ind. Specifically, the average number of input and output tokens is 23,031.39 and 2,275.24 for FB15K-237, while for Wikidata5M-Ind, the token counts are 13,099.55 and 1,254.77, respectively. Based on our empirical evaluation, the average expenditure per sample on the two datasets is approximately \$0.011 and \$0.006, respectively.

Although our inference latency and cost are marginally higher than that of lightweight knowledge graph embedding models, the fundamental advantage of SHARP lies in its plug-and-play capability.

\section{Reproducibility Map}
\label{sec:repro-map}
Table~\ref{tab:artifact-map} summarizes the artifacts used in our experiments and clarifies which parts are included in the size-limited anonymous review supplement. The uploaded supplement includes the fixed evaluation splits, entity/relation mapping files, core SHARP inference code, tool wrappers, prompt/controller logic, and the 200 curated memory trajectories. Large cached prediction traces, full baseline outputs, and plotting artifacts are not included in the review upload due to supplementary file-size limits and double-blind review constraints.

\paragraph{Minimal reproduction order.} A reviewer can inspect the fixed evaluation sets, build the memory index from the provided trajectories, run the SHARP inference pipeline on the included splits, and regenerate refreshed predictions and aggregate metrics. Exact bitwise replay is not guaranteed because hosted LLM services, Wikidata, Wikipedia, and web snippets may change over time even with temperature zero. The reported tables are therefore documented by fixed inputs, protocol details, and the included core implementation, while regenerated runs should be interpreted as refreshed live-tool executions.

\begin{table*}[tbp]
\centering
\small
\setlength{\tabcolsep}{4pt}
\renewcommand{\arraystretch}{1.08}
\begin{tabularx}{\textwidth}{@{}p{2.4cm}p{4.7cm}X@{}}
\toprule
\textbf{Stage} & \textbf{Included or documented artifact} & \textbf{Supported output} \\
\midrule
Data construction & Fixed evaluation splits for seeds 2024, 2025, and 2026; entity/relation mappings; seeded construction protocol documented in the appendix & Three balanced 1,000-instance sets per benchmark (Table~\ref{tab:evaluation-sets}) \\
Memory preparation & The 200 curated trajectories; relation/category metadata; FAISS embedding/index construction code & Top-$3$ procedural demonstrations for each query \\
Tool execution & Entity/relation resolver; KG filtering logic; Wikipedia/web wrappers; live evidence retrieval code & Anti-leakage KG observations and refreshed text evidence under live retrieval \\
Agent inference & Prompt/controller logic; tool-call parser; verdict parser; $T_{\max}=10$; batch runners for both benchmarks & Regenerated per-instance predictions, action traces, and aggregate summaries \\
Baselines & Baseline protocols, evaluation settings, and comparison methodology documented in the appendix; full baseline outputs are omitted from the size-limited review supplement & Context for the main comparison, controlled RAG comparison, and ablation results \\
Analysis & Metric definitions, per-seed evaluation protocol, and error-category summaries documented in the appendix; large cached prediction traces and plotting artifacts are omitted from the review upload & Tables, figures, variation summaries, and audited case summaries \\
\bottomrule
\end{tabularx}
\caption{Reproducibility coverage of the size-limited anonymous Code and Data Supplement.}
\label{tab:artifact-map}
\end{table*}

\section{Scope and Limitations}
\label{sec:validity}

\paragraph{Evaluation scope.}
Our balanced evaluation sets use manually verified, type-constrained
negatives to reduce trivial discrimination based on entity types.
Accordingly, the reported results characterize performance under this
controlled construction protocol rather than the naturally occurring
class distribution of a deployed KG.

\paragraph{Comparison scope.}
The evaluated baselines represent different methodological paradigms and therefore do not share identical training procedures or evidence access. Among them, Standard-RAG uses the same external knowledge sources as SHARP and provides the most direct control for assessing the contribution of SHARP's orchestration strategy. The remaining baselines contextualize performance across representative paradigms rather than constitute strictly resource-equivalent comparisons.

\paragraph{Operational scope.}
The reported results reflect the hosted models and external knowledge sources available during evaluation. Because these services may evolve, later executions may exhibit some variation even with temperature zero and fixed evaluation-set seeds. The binary decision protocol is used for benchmark evaluation only; in deployment, insufficient or conflicting evidence should trigger abstention and curator review rather than an automatic KG update.

\section{Memory Bank Construction and Quality Control}

The memory bank used by SHARP contains 200 verification trajectories covering 50 high-frequency Wikidata relations across six schema categories: Classification, Human, Geography, Time, Culture, and Organization. The goal of the memory bank is to provide reusable verification patterns rather than task-specific answers. Each trajectory records a complete Think-Act-Observe verification process, including the target triple, tool calls, returned observations, intermediate reasoning steps, and final verdict.

We construct the memory bank in three stages. First, we sample representative triples from high-frequency relations while maintaining coverage across the six schema categories. Second, an initial agent is used to generate candidate verification trajectories with the same tool interface as SHARP. Third, the generated trajectories are manually refined to improve reasoning format, remove unsupported statements, correct invalid tool-use patterns, and ensure consistency between the final verdict and the cited observations.

Manual refinement is restricted to trajectory quality control. It does not modify benchmark labels, introduce target-test answers, or add evidence unavailable to the toolset. The refinement process focuses on four aspects: (1) whether the trajectory follows the required Think-Act-Observe format; (2) whether tool calls are appropriate for the relation schema; (3) whether observations are used consistently in the final explanation; and (4) whether unsupported or hallucinated evidence is removed. Trajectories that cannot be justified by the available KG or textual evidence are discarded rather than repaired with external assumptions.

To reduce leakage risk, memory trajectories are constructed independently from the sampled test instances. During inference, the memory module retrieves only the top-k most semantically similar trajectories according to the target triple and relation semantics. The retrieved trajectories serve as procedural demonstrations for planning and tool selection; they do not provide the label of the target triple and do not expose the target triple itself. In all main experiments, we use $k=3$ retrieved trajectories.

The memory bank is therefore best understood as a small set of manually curated verification demonstrations. It provides task-level procedural guidance, such as checking entity definitions before relation-specific evidence or using external evidence when KG paths are sparse, rather than a supervised training set for the evaluated triples.

\section{Memory-Augmented Trajectory Case}
\label{sec:memory}
To demonstrate the quality of the expert knowledge stored in our Memory Bank, we present a complete reasoning trajectory in Figure~\ref{fig:memory_case}. This example illustrates how SHARP validates the triple \textbf{(Elon Musk, CEO, Tesla)}.
The trajectory showcases the agent's systematic approach: 
(1) \textbf{Schema Grounding}: Verifying the precise definitions of entities and relations in the KG (Steps 1 \& 2);
(2) \textbf{Structural Verification}: Exploring internal connectivity within the KG (Step 3);
(3) \textbf{External Cross-Validation}: Seeking real-world evidence via web search when internal paths are implicit (Step 4).
Such curated trajectories serve as procedural demonstrations for handling similar queries; they provide a tool-use pattern rather than the answer to the evaluated target.

\section{Case Study and Error Analysis}
\label{sec:case_error_analysis}

To better understand SHARP's reasoning behavior, we analyze two representative trajectories and further categorize all incorrect predictions on the two benchmarks. The cases illustrate how SHARP combines internal KG structures with external semantic evidence, while the aggregate error statistics reveal the main remaining failure modes.

\begin{table}[t]
\centering
\small
\setlength{\tabcolsep}{5pt}
\renewcommand{\arraystretch}{1.05}

\begin{tabular}{@{}lcc@{}}
\toprule
\textbf{Error Category} & \textbf{FB15K-237} & \textbf{Wikidata5M-Ind} \\
\midrule
Retrieval failure        & 37 & 10 \\
Relation granularity     & 25 & 11 \\
Plausibility bias        & 18 & 4  \\
Entity grounding         & 12 & 14 \\
Temporal/context mismatch & 6  & 13 \\
KG--text mismatch        & 11 & 5  \\
Other causes             & 19 & 6  \\
\midrule
\textbf{Total}           & \textbf{128} & \textbf{63} \\
\bottomrule
\end{tabular}
\vspace{2pt}
\caption*{\footnotesize
\textit{Note.} ``Retrieval failure'' denotes evidence sparsity or missed retrieval; 
``Relation granularity'' denotes over-rejection caused by schema or relation-scope mismatch; 
``Plausibility bias'' denotes plausibility-driven over-acceptance; 
``Entity grounding'' includes alias and identifier errors; 
``KG--text mismatch'' denotes conflicts between KG labels and external semantics.
}
\caption{Error categories for SHARP on one representative test split (seed 2026). Categories are mutually exclusive and counts sum to all incorrect predictions under this analysis protocol.}

\label{tab:error_type_distribution}
\end{table}

We categorize all incorrect predictions on the seed-2026 test set of each benchmark. The category counts sum to 128 and 63. Table~\ref{tab:error_type_distribution} summarizes the error distribution of SHARP. On FB15K-237, most errors come from evidence sparsity, retrieval failures, and relation-granularity mismatches, reflecting the difficulty of verifying fine-grained Freebase relations when supporting evidence is incomplete. On Wikidata5M-Ind, errors are more often caused by entity grounding and temporal or contextual mismatches, which is consistent with its sparse and long-tail inductive setting. These results suggest that future improvements should focus on more robust entity linking, temporal normalization, and relation-specific calibration.

\paragraph{Success Case: Biological Kingdom Boundary.}
As shown in Figure~\ref{fig:case_success_taxon}, SHARP correctly rejects a triple claiming that Kemp's Thicket Rat belongs to the plant genus \textit{Gliricidia}. The agent first identifies the strong ontological conflict between the two entities: Kemp's Thicket Rat is a mammal in Animalia, whereas \textit{Gliricidia} is a plant genus in Plantae. It then checks KG structural evidence and external semantic evidence, finding no valid support for the claimed relation. This case shows that SHARP can use high-level taxonomic constraints to produce an interpretable rejection rather than relying only on surface-level entity similarity.

\paragraph{Failure Case: Temporal Contradiction.}
Figure~\ref{fig:case_failure_temporal} shows a failure caused by temporal mismatch. SHARP correctly identifies Dat Nguyen and his affiliation with Texas A\&M football, but it treats the temporal tail entity in the triple as a strict chronological constraint. Since Nguyen was born in 1975 while the tail entity refers to the 1930s, the agent predicts the triple as incorrect. However, the benchmark label is True, suggesting that the dataset may encode a coarser affiliation relation rather than an exact time-sliced fact. This case indicates that SHARP may over-enforce logical or temporal consistency when the benchmark relation uses a coarser temporal scope than the agent infers from the tail-entity surface form.

\begin{figure*}[htbp]
    \centering
    \begin{tcolorbox}[colback=gray!5!white,colframe=gray!70!black,halign title=center, center title,title=\textbf{Prompt 2: Reasoning Instruction}, width=\linewidth]
    \small 
    \textbf{Role Definition} \\
    You are an expert Knowledge Graph Verification Agent. Your goal is to verify the factual correctness of a given triple...
    
    \textbf{Tool Definitions} \\
    (Note: The full tool definitions can be found in Appendix~\ref{sec:tool-api}. They are omitted here for the sake of brevity.) \\
    \textbf{Interaction Protocol (ReAct)} \\
    You must strictly follow this loop:
    \begin{itemize}
        \setlength{\itemsep}{0.8pt}
        \item \textbf{Thought:} Analyze the current observation and the Initial Plan. What evidence is missing?
        \item \textbf{Action:} Select a tool. Format: ToolName(value, ...)
        \item \textbf{STOP:} Do NOT generate the Observation! Stop immediately after the Action. Wait for system feedback.
        \item \textbf{Observation:} (The system will provide the tool output here)
    \end{itemize}
     ... (Repeat the loop) ...
     
    \textbf{Task Completion:} \\
    When sufficient evidence is gathered, output the final verdict: \\
    Final Answer: [Correct/Incorrect] Because [Detailed explanation based on evidence]
    
    \textbf{Judgment criteria:}... \\
    \textbf{User Context:} \\
    Strategic Plan: \{plan\} \\
    Related Reasoning Demonstrations: \{trajectory case\} \\
    Target Triple: \{triple\} \\
    Execution History: ...
    \end{tcolorbox}
    \caption{The prompt template for the Reasoning phase}
    \label{fig:prompt_reasoning}
\end{figure*}

\begin{figure*}[htbp]
    \centering
    \begin{tcolorbox}[colback=gray!5!white,colframe=gray!70!black,halign title=center, center title,title=\textbf{Prompt 3: Mandatory Judgment}, width=\linewidth]
    \small 
    \textbf{SYSTEM ALERT:} \\
    The verification process has reached the maximum step limit. 
    You MUST now make a final decision.
    
    \textbf{Reference Information} \\
    Your Initial Impression (Internal Knowledge): \{impression\} \\
    Original Plan: \{plan\} \\
    Execution History (Evidence Collected): \{history\}
    
    \textbf{Decision Logic}
    \begin{itemize}
        \setlength{\itemsep}{0.8pt}
        \item If the History contains clear evidence to prove or refute the triple (e.g., explicit KG paths or Web search results), prioritize the History.
        \item If the History is useless (e.g., tools failed, no data found), please make the most likely judgment of the triple based on your Initial Impression and History.
        \item Note: Even if History has no evidence to prove the triple, it does not necessarily mean that the triple is wrong. Please make the most likely judgment based on known information.
    \end{itemize}
    
    \textbf{Result MUST be strictly:} \\
    Final Answer: [Correct/Incorrect] Because [Detailed explanation based on evidence or fallback knowledge]
    
    Target Triple: \{triple\}
    \end{tcolorbox}
    \caption{The prompt template for the mandatory judgment phase}
    \label{fig:prompt_judgment}
\end{figure*}

\begin{figure*}[t]
\centering
\begin{tcolorbox}[colback=teal!5!white,colframe=teal!70!black,halign title=center,center title,title=\textbf{Memory Trajectory Case: (Elon Musk, CEO, Tesla)}, width=\textwidth]
\small 

\textbf{Target Triple:} "Elon Musk, CEO, Tesla"

\begin{itemize}[leftmargin=1em, itemsep=0.5em] 
    \setlength{\itemsep}{0.8pt}
    \item \textbf{Thought 1:} I need to first confirm the identity and type of "Elon Musk" to ensure we are referring to the correct person.
    \item \textbf{Action 1:} \textbf{KG\_Definition}(entity='Elon Musk')
    \item \textbf{Observation 1:} \\
    Entity Profile: Q317521 \\
    Label: Elon Musk; \\
    Description: businessman and entrepreneur (born 1971); \\ Type: human; \\ Aliases: Elon Reeve Musk.

    \item \textbf{Thought 2:} The entity 'Elon Musk' correctly refers to the businessman. Next, I should search for the definition of the "CEO" relationship to ensure my understanding is accurate.
    \item \textbf{Action 2:} \textbf{KG\_Definition}(relation='CEO')
    \item \textbf{Observation 2:} \\
    Label: chief executive officer; \\
    Description: highest-ranking corporate officer of an organization; \\
    Normalized benchmark orientation: person $\rightarrow$ organization. Wikidata corroboration may therefore require the inverse orientation of property P169 (organization $\rightarrow$ person).

    \item \textbf{Thought 3:} The relation 'CEO' means the highest-ranking corporate officer. Next, I will verify if there is a direct or 2-hop connection between Elon Musk and Tesla regarding the "CEO" relationship.
    \item \textbf{Action 3:} \textbf{KG\_Path}(entity\_a='Elon Musk', entity\_b='Tesla')
    \item \textbf{Observation 3:} \\
    (Elon Musk) --> [employer] --> (Tesla, Inc.) \\
    No unfiltered target-equivalent CEO edge is returned.

    \item \textbf{Thought 4:} The KG establishes identity and organizational association, but this does not by itself prove the specific CEO relation. I will seek explicit external evidence.
    \item \textbf{Action 4:} \textbf{Web\_Evidence}(question='Is Elon Musk the CEO of Tesla?')
    \item \textbf{Observation 4:} \\
    Snippet 1: "...Tesla CEO Elon Musk Gets a Long Do-Not-Tweet List From the SEC..." (Source: Barron's) \\
    Snippet 2: "...Elon Musk is the co-founder and CEO of Tesla..." (Source: Official Bio)
    ......

    \item \textbf{Thought 5:} Based on the current evidence, it is sufficient to prove the triple.
    \item \textbf{Final Answer:} [Correct] Because the entity and organization are correctly grounded, and the retrieved text explicitly identifies Elon Musk as Tesla's CEO. The employer edge is supporting context, not direct proof of the CEO relation.
\end{itemize}
\end{tcolorbox}
\caption{A representative reasoning trajectory from the Memory Bank. This case demonstrates the ``Think-Act-Observe'' loop where the agent fuses KG structural info with Web evidence.}
\label{fig:memory_case}
\end{figure*}

\begin{figure*}[t]
\centering
\begin{tcolorbox}[
    colback=green!5!white, 
    colframe=green!60!black, 
    halign title=center, 
    center title, 
    title=\textbf{Successful Verification Trajectory}, 
    width=\textwidth
]
\small 

\textbf{Target Triple:} (kemp's thicket rat, parent taxon, gliricidia) \quad \textbf{Ground Truth:} False \quad \textbf{Verdict:} [Incorrect]

\begin{itemize}[leftmargin=1.5em, itemsep=0.3em]
    \item \textbf{Thought 1:} I need to first confirm the biological classification and taxonomic rank of "kemp's thicket rat" to understand what kind of entity it is and whether it could logically have a "parent taxon" relationship with "gliricidia".
    \item \textbf{Action 1:} \textbf{KG\_Definition}(entity: "kemp's thicket rat")
    \item \textbf{Observation 1:} Label: Kemp's Thicket Rat; Description: species of mammal; Type: taxon; Aliases: Thamnomys kempi.
    
    \item \textbf{Thought 2:} The entity "Kemp's Thicket Rat" is confirmed as a taxon and specifically a species of mammal. Next, I need to verify what "Gliricidia" refers to.
    \item \textbf{Action 2:} \textbf{KG\_Definition}(entity: "gliricidia")
    \item \textbf{Observation 2:} Label: Gliricidia; Description: genus of plants; Type: taxon.
    
    \item \textbf{Thought 3:} Both are confirmed as taxa, but they belong to entirely different biological kingdoms: Animalia (mammal) vs. Plantae (plant genus). A parent taxon relationship cannot cross kingdom boundaries. I will check for any structural paths.
    \item \textbf{Action 3:} \textbf{KG\_Path}(entity\_a: "kemp's thicket rat", entity\_b: "gliricidia")
    \item \textbf{Observation 3:} No direct, 2-hop or 3-hop paths found between Q1761125 and Q12549487.
    
    \item \textbf{Thought 4:} The knowledge graph shows no structural paths, which aligns with biological expectations. I will perform a final web search to rule out any obscure claims.
    \item \textbf{Action 4:} \textbf{Web\_Evidence}(question: "Is Kemp's thicket rat taxonomically related to Gliricidia?")
    \item \textbf{Observation 4:} No relevant webpages found (API returned empty).
    
    \item \textbf{Final Answer:} [Incorrect] Because "Kemp's thicket rat" is a mammalian species (Animalia), while "Gliricidia" is a genus of plants (Plantae). The "parent taxon" relationship strictly applies within a lineage and cannot cross kingdom boundaries. The KG confirms no structural path exists, consistent with biological classification principles.
\end{itemize}
\end{tcolorbox}
\caption{Success Case: Cross-Kingdom Ontological Reasoning}
\label{fig:case_success_taxon}
\end{figure*}

\begin{figure*}[t]
\centering
\begin{tcolorbox}[
    colback=red!5!white, 
    colframe=red!60!black, 
    halign title=center, 
    center title, 
    title=\textbf{Error Verification Trajectory}, 
    width=\textwidth
]
\small 

\textbf{Target Triple:} (Nguy$\tilde{\text{e}}$n T$\acute{\text{a}}$n Đ$\d{a}$t, member of sports team, texas a\&m aggies football(1930-1939)) \quad \\ \textbf{Ground Truth:} True \quad \textbf{Agent Prediction:} [Incorrect] \textbf{(Error)}

\begin{itemize}[leftmargin=1.5em, itemsep=0.3em]
    \item \textbf{Thought 1:} I need to first verify the identity and background of "Nguyen Tan Dat" to determine if this person could plausibly have been associated with a U.S. college football team in the 1930s.
    \item \textbf{Action 1:} \textbf{KG\_Definition}(entity: "Nguyen Tan Dat")
    \item \textbf{Observation 1:} Label: Dat Nguyen; Description: All-American college football player; Aliases: Dat Tan Nguyen.
    
    \item \textbf{Thought 2:} This refers to Dat Nguyen, a modern player. I need to check the team entity "texas a\&m aggies football, 1930-1939" to see if it exists as a distinct time-sliced entity.
    \item \textbf{Action 2:} \textbf{KG\_Definition}(entity: "texas a\&m aggies football, 1930-1939")
    \item \textbf{Observation 2:} Label: Texas A\&M Aggies football; Description: University American football team.
    
    \item \textbf{Thought 3:} The KG only contains the general team entity. I will check the player's birth date and team membership to verify the 1930s timeframe.
    \item \textbf{Action 3:} \textbf{KG\_Neighbor}(entity: "Nguyen Tan Dat", relation: "member of sports team")
    \item \textbf{Observation 3:} [member of sports team] -> Texas A\&M Aggies football; [date of birth] -> 1975-09-25.
    
    \item \textbf{Thought 4:} Dat Nguyen was born in 1975, making it chronologically impossible for him to have played in the 1930s. I will perform a final web search for any historical namesake.
    \item \textbf{Action 4:} \textbf{Web\_Evidence}(question: "Was there a Vietnamese player named Dat Nguyen on Texas A\&M in the 1930s?")
    \item \textbf{Observation 4:} No relevant webpages found (API returned empty).
    
    \item \textbf{Final Answer:} [Incorrect] Because Nguyen Tan Dat (Dat Nguyen) was born in 1975 and played for Texas A\&M in the 1990s, making it chronologically impossible for him to be a member between 1930 and 1939.
\end{itemize}
\end{tcolorbox}
\caption{Failure Case: Chronological Impossibility and Data Noise}
\label{fig:case_failure_temporal}
\end{figure*}
\end{document}